\documentclass[10pt,letterpaper]{article}
\usepackage[top=0.85in,left=2.75in,footskip=0.75in]{geometry}

\usepackage{amsmath,amssymb}

\usepackage{changepage}

\usepackage{textcomp,marvosym}
\usepackage[utf8]{inputenc}
\usepackage[T1]{fontenc}
\usepackage{float}
\usepackage{algorithm}
\usepackage{algpseudocode}
\usepackage{float}
\usepackage{tabularx}
\usepackage{booktabs}
\usepackage{cite}

\usepackage{nameref,hyperref}

\usepackage[right]{lineno}

\usepackage[nopatch=eqnum]{microtype}
\DisableLigatures[f]{encoding = *, family = * }

\usepackage[table]{xcolor}

\usepackage{array}

\newcolumntype{+}{!{\vrule width 2pt}}

\newlength\savedwidth

\raggedright
\usepackage[aboveskip=1pt,labelfont=bf,labelsep=period,justification=raggedright,singlelinecheck=off]{caption}

\makeatletter
\renewcommand{\@biblabel}[1]{\quad#1.}
\makeatother

\usepackage{lastpage,fancyhdr,graphicx}
\usepackage{epstopdf}
\fancyheadoffset[L]{2.25in}
\fancyfootoffset[L]{2.25in}
\begin{document}
\vspace*{0.2in}

\begin{flushleft}
{\Large
\textbf\newline{Prediction of Significant Creatinine Elevation in First ICU Stays with Vancomycin Use: A retrospective study through Catboost} % Please use "sentence case" for title and headings (capitalize only the first word in a title (or heading), the first word in a subtitle (or subheading), and any proper nouns).
}
\newline
\\
Junyi Fan\textsuperscript{1},
Li Sun\textsuperscript{1},
Shuheng Chen\textsuperscript{1},
Yong Si\textsuperscript{1},
Minoo Ahmadi\textsuperscript{1},
Greg Placencia\textsuperscript{2},
Elham Pishgar\textsuperscript{3},
Kamiar Alaei\textsuperscript{4},
Maryam Pishgar\textsuperscript{1,*}
\\
\bigskip
\textbf{1} Department ofIndustrial and Systems Engineering, University of Southern California, Los Angeles, California, United States
\\
\textbf{2} Department of Industrial and Manufacturing Engineering, California State Polytechnic University Pomona, Pomona, California, United States
\\
\textbf{3} Colorectal Research Center, lran University of Medical Sciences, Tehran, Iran
\\
\textbf{4} Department of Health Science, California State University Long Beach, Long Beach, California, United States
\\
\bigskip

% Insert additional author notes using the symbols described below. Insert symbol callouts after author names as necessary.
% 
% Remove or comment out the author notes below if they aren't used.
%
% Primary Equal Contribution Note

% Use the asterisk to denote corresponding authorship and provide email address in note below.
* pishgar@usc.edu

\end{flushleft} 
\section*{Abstract}
\textbf{Background:} Vancomycin, a critical antibiotic for treating severe Gram-positive infections in intensive care units, carries substantial nephrotoxic risk. Despite advances in understanding vancomycin-induced nephrotoxicity mechanisms, early prediction of renal injury remains challenging in critically ill patients. This study aimed to develop a machine learning framework for predicting vancomycin-associated creatinine elevation using routinely collected clinical data.

\textbf{Methods:} We conducted a retrospective analysis using MIMIC-IV, identifying 10,288 ICU patients aged 18-80 years who received vancomycin. The primary outcome was vancomycin-associated renal injury, defined by KDIGO criteria as creatinine elevation $\geq$ 0.3 mg/dL within 48 hours or $\geq$ 50\% increase within 7 days. Feature selection used a two-stage approach: SelectKBest with F-statistics identified the top 30 features, then Random Forest importance ranking selected the final 15 variables. Six machine learning algorithms were evaluated using 5-fold cross-validation, with interpretability assessed through SHAP analysis, Accumulated Local Effects, and Bayesian posterior sampling.

\textbf{Results:} Among 10,288 patients, 2,903 (28.2\%) developed vancomycin-associated creatinine elevation. CatBoost achieved the highest predictive performance with AUROC of 0.818 (95\% CI: 0.801–0.834), sensitivity of 0.800, specificity of 0.681, and negative predictive value of 0.900. Key predictors included phosphate, total bilirubin, magnesium, Charlson comorbidity index, and APSIII score. SHAP analysis confirmed that elevated phosphate levels substantially increased predicted risk, while ALE plots revealed clinically interpretable dose-response relationships. Bayesian uncertainty quantification demonstrated mean risk prediction of 60.5\% with 95\% credible interval of 16.8\%-89.4\% for high-risk patients.

\textbf{Conclusions:} This machine learning framework successfully predicts vancomycin-associated creatinine elevation using routinely collected ICU data, achieving robust discriminatory performance while maintaining clinical interpretability. The model identifies patients at high risk for renal injury before clinically apparent kidney dysfunction, potentially enabling timely interventions and improved antimicrobial stewardship in critical care settings.

% \linenumbers

% Use "Eq" instead of "Equation" for equation citations.
\section*{Introduction}
Vancomycin, a glycopeptide antibiotic first isolated from Amycolatopsis orientalis in the 1950s, remains one of the most important treatments for severe infections caused by Gram-positive pathogens, particularly methicillin-resistant Staphylococcus aureus (MRSA) and Enterococcus species\cite{griffith1981introduction, levine2006vancomycin, moellering2006vancomycin}. Its bactericidal activity stems from its ability to inhibit bacterial cell wall synthesis by binding to the D-Ala-D-Ala terminus of peptidoglycan precursors, thus preventing cell wall formation\cite{liu2011clinical}. Despite the emergence of vancomycin-resistant Enterococcus (VRE) and other resistant strains, vancomycin continues to be a cornerstone in the treatment of serious hospital-acquired infections, especially in settings where multidrug resistance is prevalent\cite{stevens2006role,he2020evidence,kirst1998historical}.

While essential in treating severe infections, vancomycin is associated with several side effects, the most concerning of which is nephrotoxicity. Multiple in vivo and in vitro studies have demonstrated that high intracellular concentrations lead to oxidative stress, mitochondrial dysfunction, and resultant proximal tubular apoptosis and necrosis\cite{kwiatkowska2021mechanism,campbell2023overview,kan2022vancomycin,dvzidic2024unveiling}. Lysosomal accumulation of the drug also contributes to lysosomal–mitochondrial crosstalk and cell death . Moreover, allergic mechanisms such as acute interstitial nephritis, and tubular cast formation further mediate renal injury\cite{kan2022vancomycin,hammond2017systematic,venugopalan2024association}. However, despite advances in understanding these molecular pathways, the overall pathogenesis of vancomycin-induced nephrotoxicity remains incompletely understood, as significant individual variability and complex interactions between different injury mechanisms continue to be observed. 

A lot of studies have been conducted to narrow this gap. For example, Aljefri et al. (2019) performed a systematic review analyzing the relationship between vancomycin AUC and acute kidney injury, synthesizing data from eight observational studies involving 2,491 patients\cite{aljefri2019vancomycin}. Their meta-analysis demonstrated that maintaining vancomycin AUC below approximately 650 mg·h/L within the first 48 hours was associated with a significantly lower risk of AKI, and that AUC-guided monitoring strategies resulted in less nephrotoxicity compared to traditional trough-guided approaches, highlighting the importance of pharmacokinetic monitoring in reducing vancomycin-associated kidney injury. Kim et al. (2022) constructed a risk scoring system for vancomycin-associated acute kidney injury by retrospectively analyzing clinical data with logistic regression and machine learning methods, including random forest and support vector machine, achieving AUROC values between 0.72 and 0.74\cite{kim2022risk}. Their study identified low bodyweight, higher Charlson comorbidity index, elevated vancomycin trough levels, and concomitant use of multiple nephrotoxic agents as important predictors, providing a practical tool for risk assessment in clinical settings.

Given the complex and multifactorial mechanisms underlying vancomycin-induced nephrotoxicity, there is an urgent need for more precise clinical risk stratification—especially in critically ill ICU patients who are inherently more vulnerable to kidney injury. Serum creatinine elevation is a key indicator of kidney injury and is closely monitored during vancomycin therapy. By leveraging machine learning approaches to analyze the relationship between vancomycin administration and creatinine dynamics, our study aims to identify which variables most strongly predict creatinine elevation and subsequent renal complications in ICU populations. This approach not only enables the identification of high-risk patients and modifiable risk factors (such as drug exposure levels, concomitant medications, and patient-specific characteristics) but also provides a data-driven foundation for optimizing vancomycin use. Ultimately, these insights can improve individualized dosing strategies, minimize the incidence of vancomycin-associated kidney injury, and enhance patient safety in the ICU setting.

This study makes several significant methodological and clinical contributions to vancomycin nephrotoxicity prediction in critical care settings:

\begin{itemize}
   \item \textbf{Temporal Precision in Outcome Definition}: Unlike previous studies using pre-existing AKI diagnosis labels, we employed time-stamped serum creatinine trajectories to define vancomycin-associated renal injury. This addresses critical data leakage where AKI diagnoses lack temporal precision, making it impossible to confirm kidney injury occurred \textit{after} vancomycin administration. Our KDIGO-based criteria ($\geq$0.3 mg/dL within 48 hours or $\geq$50\% increase within 7 days post-vancomycin) ensure strict temporal validity and eliminate reverse causality—where pre-existing renal dysfunction is misclassified as drug-induced injury.

   \item \textbf{Comprehensive Feature Engineering with Clinical Validation}: We developed a multi-domain feature extraction pipeline capturing clinical state immediately prior to vancomycin across three data sources: real-time monitoring (\texttt{chartevents}), laboratory evaluation (\texttt{labevents}), and procedural interventions (\texttt{procedureevents}). Each feature represents the latest measurement before drug initiation, mirroring real-world clinical scenarios while maintaining strict temporal boundaries to prevent data leakage and enhance deployability.

   \item \textbf{Posterior-Based Risk Stratification with Uncertainty Quantification}: We leveraged DREAM algorithm to estimate full nephrotoxicity risk distributions rather than point predictions. By conditioning on priors from patients with creatinine elevation, DREAM outputs individualized probability curves with explicit uncertainty bounds. This allows clinicians to assess both risk level and prediction confidence, supporting threshold-free decisions where wide confidence intervals suggest monitoring while narrow intervals support definitive interventions.

   \item \textbf{Systematic Model Benchmarking with Clinical Interpretability}: We evaluated six diverse machine learning paradigms spanning linear models, ensemble methods, and neural networks, identifying CatBoost as optimal. Multiple interpretability techniques—SHAP analysis, Accumulated Local Effects, and ablation studies—create a transparent, clinically auditable framework supporting evidence-based bedside decision-making through physiologically plausible predictors.
\end{itemize}

\section*{Materials and Methods}
\subsection*{Data Source and study design}

The study was conducted retrospectively using MIMIC-IV (v2.2)\cite{johnson2020mimic}, a large-scale, de-identified database of ICU admissions curated by the Massachusetts Institute of Technology and Beth Israel Deaconess Medical Center. MIMIC-IV contains comprehensive, time-stamped data on demographics, vital signs, laboratory tests, medication exposures, and procedures for ICU patients admitted between 2008 and 2019. Leveraging this database, we constructed a binary classification framework to predict renal injury risk using only clinical information available prior to the first vancomycin dose. The entire experimental pipeline—including data preprocessing, feature selection, model development, performance evaluation, and explainability—was grounded in clinical relevance and designed for translational applicability to ICU workflows.

To operationalize the entire machine learning pipeline, we constructed a modular, step-wise framework encompassing cohort definition, data preprocessing, feature selection, model training, and interpretation. The full workflow is outlined below in pseudocode format:

\begin{algorithm}[H]
\caption{Machine Learning Framework for Predicting Vancomycin-Associated Renal Injury}
\begin{algorithmic}[1]
\Require MIMIC-IV ICU data (2008–2019)
\Ensure Binary prediction of creatinine elevation risk following vancomycin exposure
\State \textbf{Step 1: Study Cohort Construction}
\State Identify adult ICU patients who received IV vancomycin
\State Apply inclusion criteria: age between 18 and 80, no active cancer
\State Retain first ICU stay only
\State Define outcome using KDIGO-based thresholds: $\Delta$Cr $\geq$ 0.3 mg/dL within 48h or $\geq$ 50\% rise within 7 days

\State \textbf{Step 2: Data Preprocessing}
\State Extract most recent values prior to vancomycin initiation
\State Impute continuous variables with median; categorical with mode
\State Normalize continuous variables using min-max scaling
\State Retain binary indicators as 0/1

\State \textbf{Step 3: Feature Selection}
\State Group features by source (chart events, lab events, procedure events)
\State Perform univariate statistical testing via SelectKBest with F-statistics ($p > 0.05$)
\State Select top 30 features with highest F-statistics from univariate analysis
\State Apply Random Forest importance ranking to identify final 15 most predictive variables
\State Combine selected features with admission-level characteristics

\State \textbf{Step 4: Imbalance Handling}
\State Apply SMOTE oversampling within training folds to address class imbalance

\State \textbf{Step 5: Model Development}
\ForAll{models $\in$ \{CatBoost, LightGBM, XGBoost, Logistic Regression, Naïve Bayes, Shallow Neural Net\}}
    \State Perform stratified 5-fold cross-validation
    \State Optimize hyperparameters via grid search
    \State Evaluate AUROC, F1-score, sensitivity, specificity, PPV, NPV
\EndFor

\State \textbf{Step 6: Posthoc Analysis and Interpretation}
\State Assess statistical equivalence of training/test sets via t-tests
\State Quantify marginal utility via ablation analysis
\State Interpret global/local effects using SHAP and ALE
\State Quantify predictive uncertainty via DREAM posterior sampling
\end{algorithmic}
\end{algorithm}

This structured design ensured that each methodological step—from cohort assembly to final interpretability—was aligned with clinical reasoning, statistically rigorous, and transparent for deployment in real-world ICU settings.

\subsection*{Study Population}
We retrospectively identified a cohort of ICU patients who received intravenous vancomycin to investigate risk factors associated with drug-induced renal injury. A stepwise screening process was applied to ensure clinical relevance and to minimize potential confounding.

The initial cohort included 25,916 ICU stays involving adult patients who had received at least one dose of intravenous vancomycin. This broad inclusion criterion ensured that all relevant exposures were captured. To reduce physiological heterogeneity and age-related renal confounding, we restricted the cohort to patients aged between 18 and 80 years, excluding those outside this range. Pediatric patients differ in renal maturity, pharmacokinetics, and treatment protocols, while elderly patients often have baseline renal impairment, increasing the likelihood of non–drug-related creatinine fluctuations. This step resulted in 21,925 eligible patients.

Next, we excluded patients with active malignancy or metastatic cancer, as these conditions may independently affect renal function through paraneoplastic processes, chemotherapy, or end-of-life care, which could confound the renal effects attributed to vancomycin. After removing 2,720 such cases, 19,205 patients remained. To avoid duplicate representations and ensure that extracted features reflected the initial stage of critical illness, we retained only the first ICU admission per patient. This yielded a final study population of 10,288 unique ICU stays.

 The prediction target was defined based on clinical consensus criteria from the Kidney Disease: Improving Global Outcomes (KDIGO) guidelines and widely adopted nephrotoxicity thresholds\cite{khwaja2012kdigo,kellum2013diagnosis}. Specifically, a patient was considered to have vancomycin-associated renal injury if either of the following criteria were met after the initial vancomycin dose: (1) an absolute increase in serum creatinine of at least 0.3 mg/dL within 48 hours, or (2) a relative increase of 50\% or more in peak creatinine within 7 days compared to the most recent pre-vancomycin baseline. These criteria reflect clinically meaningful renal impairment and align with pharmacovigilance standards for nephrotoxic agents.A total of 2,903 patients met at least one of these criteria and were labeled positive, while the remaining 7,385 patients served as the negative group.

\begin{figure}[H]
\centering
\includegraphics[width=0.85\linewidth]{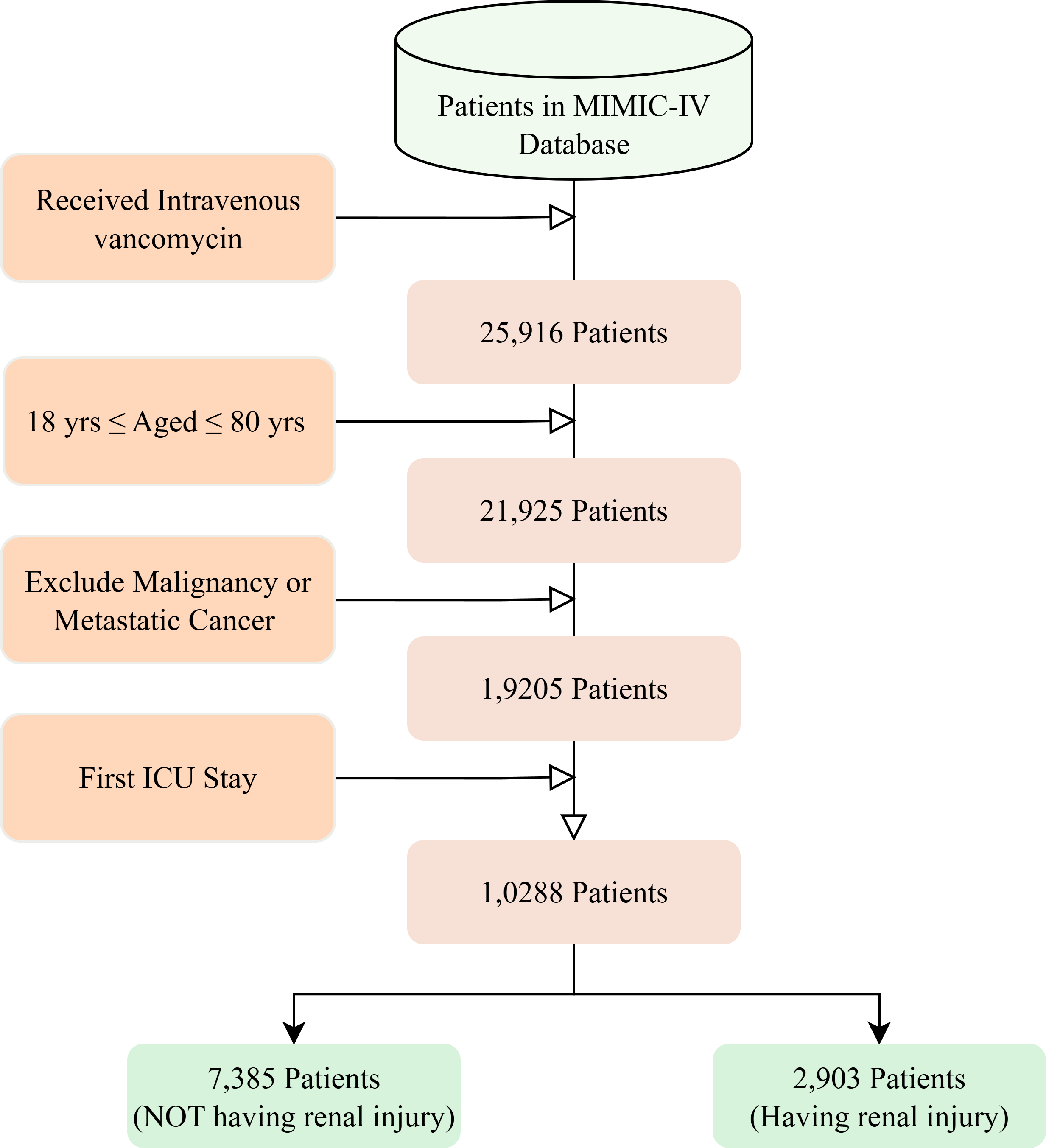}
\caption{Cohort selection process for vancomycin-associated renal injury analysis}
\label{fig:patient_selection}
\end{figure}
\subsection*{Data Preprocessing}

In this study, we developed a preprocessing pipeline specifically tailored to the structure of ICU data and the clinical context of vancomycin use. All predictor variables were extracted based on the most recent available measurement prior to the first vancomycin dose, ensuring strict temporal validity for prospective risk modeling.

We applied different imputation strategies to continuous and categorical variables. For continuous variables, including laboratory and physiological measures such as \textit{Phosphate, Anion Gap},  missing values were imputed using the median from the training set\cite{fan2025predictingshorttermmortalityelderly}. This method is robust to outliers and skewed distributions, which are common in critical care laboratory data. For categorical variables such as \textit{Richmond-RAS Scale}, \textit{Braden Mobility}, and presence of an \textit{Arterial Line}, mode imputation was used to preserve the most frequent clinical state and reduce noise from documentation inconsistencies.

Because our study aimed to predict renal outcomes based on patient status at the time of drug initiation, we did not compute temporal summary statistics (e.g., max, min, or mean over 24 hours). Instead, each variable was represented by a single value—the latest recorded measurement prior to vancomycin administration. This approach mirrors clinical practice and enhances the interpretability of predictions in real-time applications.

To ensure scale comparability across variables, min-max normalization was applied to all continuous variables. This transformation mapped raw values to the $[0,1]$ interval and prevented features with large numerical ranges, such as \textit{lactate} or \textit{platelet count}, from dominating learning algorithms. Binary variables, including indicator flags for devices, procedures, and comorbidities, were retained in their native 0/1 form to preserve clinical meaning and compatibility with tree-based and linear models.

The dataset exhibited a moderate class imbalance, with substantially fewer patients meeting the criteria for vancomycin-associated renal injury. To address this, we applied the Synthetic Minority Over-sampling Technique to the training folds during cross-validation\cite{chen2025interpretablemachinelearningmodel}. SMOTE generated synthetic minority-class samples by interpolating between existing observations, helping the model learn boundary regions more effectively while avoiding overfitting. Specifically, for a minority-class instance $x_{\text{minority}}$ and one of its $k$-nearest neighbors $x_{\text{neighbor}}$, a synthetic sample $x_{\text{new}}$ is created as:

\begin{equation}
x_{\text{new}} = x_{\text{minority}} + \delta \cdot (x_{\text{neighbor}} - x_{\text{minority}}), \quad \delta \sim U(0,1)
\end{equation}

where $\delta$ is a random scalar drawn from a uniform distribution. This oversampling step was restricted to the training data and excluded from test folds to ensure unbiased evaluation.

All preprocessing steps—including imputation, normalization, encoding, and oversampling—were conducted independently on training data and subsequently applied to the corresponding validation sets. This design avoided information leakage and ensured that model evaluation reflected performance on unseen data. The resulting feature matrix provided a temporally aligned, numerically stable, and clinically interpretable basis for predictive modeling.

\subsection*{Feature Selection}
We initially compiled a comprehensive set of candidate predictors from MIMIC-IV's three primary clinical data tables, each representing distinct aspects of ICU patient monitoring and care. This systematic approach ensured comprehensive coverage of physiological systems associated with vancomycin-induced nephrotoxicity while maintaining clinical interpretability.

The \texttt{chartevents} table contains real-time physiological measurements and clinical assessments documented at the bedside, representing continuous patient monitoring data. This table captures vital signs, neurological assessments, and point-of-care laboratory values that reflect immediate patient status. From this domain, we extracted features including \textit{Richmond-RAS Scale, Total Bilirubin, Arterial Base Excess, AST, Braden Mobility, Mean Airway Pressure}, along with additional real-time observations such as \textit{Heart Rate, Non-Invasive Blood Pressure}, and \textit{SpO\textsubscript{2}} that were initially considered. These variables represent the dynamic physiological state and acute illness severity that may predispose patients to vancomycin nephrotoxicity.

The \texttt{labevents} table encompasses formal laboratory test results processed in hospital laboratories, providing biochemical markers of organ function and metabolic status. This systematic laboratory evaluation offers objective measures of renal function, electrolyte balance, coagulation status, and metabolic derangements. Key features from this domain included \textit{Phosphate, Anion Gap, Magnesium, Lactate, PTT, Platelet Count, White Blood Cells}, and \textit{Glucose}, as well as additional candidates like \textit{BUN, INR}, and \textit{Calcium} that were reviewed during initial screening \cite{du2020new}. These laboratory parameters are critical for assessing baseline renal vulnerability and identifying patients at higher risk for drug-induced nephrotoxicity.

The \texttt{procedureevents} table documents invasive procedures and interventions performed during ICU care, reflecting both illness severity and therapeutic intensity. Procedural data indicates the level of medical intervention required and serves as a proxy for critical illness severity. From this domain, we included features such as the presence of an \textit{Arterial Line, central line insertion} and \textit{mechanical ventilation initiation} \cite{soenksen2022integrated}. These interventional markers provide insights into hemodynamic monitoring needs, vascular access requirements, and respiratory support intensity, all of which correlate with nephrotoxicity risk in critically ill patients receiving vancomycin.

This clinically grounded three-domain approach reflects the fundamental pillars of ICU care—continuous physiologic monitoring, systematic biochemical evaluation, and therapeutic intervention intensity—enabling structured feature selection while preserving alignment with clinical decision-making processes for renal risk assessment.

Our feature selection employed a two-stage approach combining statistical filtering with machine learning-based importance ranking to identify the most predictive variables for vancomycin-associated nephrotoxicity while minimizing overfitting and maintaining clinical interpretability.

\textbf{Stage 1: Statistical Filtering (SelectKBest)}
The initial stage utilized univariate statistical testing to reduce dimensionality and eliminate non-informative features. We applied the F-statistic from ANOVA F-test via SelectKBest with f\_classif scoring function to evaluate each feature's individual association with nephrotoxicity outcome \cite{pedregosa2011scikit}:

\begin{equation}
F = \frac{\text{MS}_{\text{between}}}{\text{MS}_{\text{within}}} = \frac{\sum_{i} n_i (\bar{X}_i - \bar{X})^2 / (k-1)}{\sum_{i,j} (X_{ij} - \bar{X}_i)^2 / (N-k)}
\end{equation}

where $k$ represents the number of groups (nephrotoxicity vs. no nephrotoxicity), $n_i$ is the sample size of group $i$, and $N$ is the total sample size. This approach selected the top 30 features with the highest F-statistics, effectively filtering variables that demonstrated significant group differences in vancomycin-treated ICU patients.

\textbf{Stage 2: Machine Learning-Based Ranking (Random Forest Importance)}
From the 30 statistically significant features, we applied Random Forest feature importance to identify the most predictive subset for nephrotoxicity risk \cite{breiman2001random}. Random Forest importance quantifies each feature's contribution to prediction accuracy by measuring the mean decrease in node impurity across all trees \cite{louppe2013understanding}:

\begin{equation}
\text{Importance}(X_j) = \frac{1}{T} \sum_{t=1}^{T} \sum_{v \in V_t} p(v) \cdot \Delta I(v, X_j)
\end{equation}

where $T$ is the number of trees, $V_t$ represents nodes in tree $t$ that split on feature $X_j$, $p(v)$ is the proportion of samples reaching node $v$, and $\Delta I(v, X_j)$ is the Gini impurity decrease from splitting on feature $X_j$ at node $v$. This method has proven effective in medical prediction tasks, particularly for identifying key risk factors in critical care settings \cite{chen2018machine}.

This two-stage approach selected the final 15 most important features, balancing statistical significance with predictive utility while reducing overfitting risk. The combination of univariate filtering and ensemble-based ranking ensures both statistical validity and clinical relevance for vancomycin nephrotoxicity prediction, following established practices in medical machine learning applications \cite{rajkomar2018scalable}.

In addition to the three clinical event categories, we incorporated four key admission-level characteristics that establish baseline patient risk profiles. \textit{Age} serves as a fundamental predictor of vancomycin nephrotoxicity due to reduced renal reserve and altered drug clearance in older patients \cite{lodise2009relationship}. \textit{Emergency Department (ED) duration} reflects clinical acuity and complexity prior to ICU admission, potentially indicating hemodynamic instability that may predispose to renal injury. \textit{Charlson Comorbidity Index} quantifies pre-existing chronic disease burden using admission diagnoses, providing standardized baseline health status assessment that influences nephrotoxicity susceptibility \cite{charlson1987new}. \textit{Acute Physiology Score III (APSIII)} measures acute illness severity during the first 24 hours of ICU admission, capturing physiological derangement that correlates with organ dysfunction risk \cite{knaus1991apache}. These admission-level features complement dynamic clinical measurements by establishing foundational risk context, enabling the model to account for both baseline vulnerability and acute physiological changes.

The final selected features are summarized in Table~\ref{tab:final_features_vanco}.

\begin{table}[H]
\centering
\caption{Final selected features used for predicting vancomycin-associated renal injury}
\label{tab:final_features_vanco}
\renewcommand{\arraystretch}{1.2}
\small
\begin{tabular}{p{3.5cm}|p{10.5cm}}
\hline
\rowcolor[HTML]{f7e1d7}
\textbf{Category} & \textbf{Selected Features} \\
\hline
\texttt{chartevents} & 
Richmond-RAS Scale, Total Bilirubin, Arterial Base Excess, AST, Braden Mobility, Mean Airway Pressure \\
\hline
\texttt{procedureevents} & 
Arterial Line \\
\hline
\texttt{labevents} & 
Phosphate, Anion Gap, Magnesium, Lactate, PTT, Platelet Count
, White Blood Cells, Glucose, APS III \\
\hline
\texttt{admission} & 
Age, ED duration, Magnesium, Charlson Comorbidity Index \\
\hline
\end{tabular}
\end{table}

\subsection*{Modeling}

To predict vancomycin-associated renal injury in ICU patients, we developed a supervised machine learning framework incorporating six representative classification algorithms, each selected for its theoretical strengths, suitability for clinical data, and complementary modeling capabilities. The dataset was randomly split into stratified training (70\%) and test (30\%) sets to preserve outcome distribution. All model development—including hyperparameter tuning and performance validation—was conducted using five-fold stratified cross-validation within the training set to avoid information leakage and ensure generalizability.

We employed random stratified splitting instead of temporal splitting based on admission dates for medical and data science considerations. From a medical perspective, vancomycin nephrotoxicity mechanisms involve fundamental physiological processes -oxidative stress, tubular injury, and mitochondrial dysfunction - that remain biologically consistent over time, making temporal evolution less relevant than comprehensive patient representation. From a data science perspective, random splitting maximizes model performance by incorporating the most recent clinical practices and contemporary protocols in training, while temporal splitting would force the model to learn from outdated 2008-2014 data and exclude valuable 2015-2019 insights. Our approach ensures optimal clinical applicability for current deployment while maintaining statistical rigor through robust cross-validation techniques.

We prioritized CatBoost, LightGBM, and XGBoost as core modeling algorithms due to their proven effectiveness in handling structured clinical datasets. These tree-based ensemble methods leverage gradient boosting to sequentially improve prediction accuracy while capturing non-linear feature interactions and hierarchical patterns. CatBoost was particularly advantageous for its ordered boosting strategy and native support for categorical variables, reducing overfitting risks without extensive preprocessing. LightGBM offered efficiency through histogram-based binning and leaf-wise growth, which accelerated training on high-dimensional data. XGBoost provided granular control over regularization, tree complexity, and sampling strategies, allowing for flexible bias–variance trade-offs.

For interpretability, we included logistic regression with both L1 (lasso) and L2 (ridge) penalties to serve as a transparent baseline model. Its linear structure allowed for direct inspection of coefficient weights, enabling clinicians to interpret variable effects in a familiar framework. Regularization was applied to control multicollinearity and overfitting, with hyperparameters optimized via grid search.

A Gaussian Naïve Bayes classifier was also evaluated as a probabilistic baseline. This lightweight model assumes conditional independence among predictors and models continuous variables using parametric likelihoods. While simplistic, its efficiency and interpretability make it a useful benchmark for gauging the added value of more expressive algorithms.

Lastly, we implemented a shallow feedforward neural network to explore non-linear, high-capacity representations. The architecture included a single hidden layer with ReLU activation and a sigmoid output node. Training was performed using the Adam optimizer and binary cross-entropy loss, with hyperparameters such as learning rate, dropout ratio, and batch size selected via nested tuning. Although more data-intensive and less interpretable, neural networks offer flexible function approximation that may be valuable in larger-scale or multimodal extensions of this work.

Model performance was primarily evaluated using the area under the receiver operating characteristic curve (AUROC), which provides a threshold-independent measure of discrimination. Given the class imbalance in renal injury outcomes, AUROC served as a robust metric for comparing overall model performance. To quantify variability and assess stability, 95\% confidence intervals for AUROC were calculated using 2{,}000 bootstrap replicates.

In addition to AUROC, we reported a set of complementary metrics to ensure clinical applicability. Accuracy provided a general overview of classification performance, while the F1-score balanced precision and recall—important for minimizing both false positives and false negatives. Sensitivity and specificity were included to evaluate the model’s ability to detect high-risk cases and avoid overtreatment, respectively. Positive predictive value (PPV) and negative predictive value (NPV) were also calculated to aid in clinical interpretation of the model's predictions. This comprehensive evaluation strategy ensured a balanced assessment of both statistical performance and real-world utility.

Together, these six models span a diverse spectrum of complexity, interpretability, and inductive bias, allowing for a comprehensive assessment of machine learning paradigms in predicting early renal complications related to vancomycin administration.

\subsection*{Statistical Analyses}

To ensure the clinical relevance, interpretability, and methodological robustness of our vancomycin-associated renal outcome prediction framework, we implemented a suite of statistical analyses tailored to five key objectives: (1) to validate the comparability of training and test cohorts; (2) to quantify the marginal contribution of each feature through ablation; (3) to uncover non-linear or threshold effects using Accumulated Local Effects; (4) to interpret individual-level predictions via SHAP; and (5) to estimate predictive uncertainty through posterior sampling. These analyses were designed to support both the statistical credibility and bedside applicability of our model for predicting early renal injury and creatinine elevation after vancomycin exposure.

We first evaluated the statistical equivalence of the training and test sets using two-sided independent-sample t-tests across core clinical variables\cite{starsurg2018association} including laboratory results and physiologic scores. Welch’s correction was applied when unequal variances were detected. The t-statistic was computed as:

\begin{equation}
t = \frac{\bar{x}_1 - \bar{x}_2}{\sqrt{\frac{s_1^2}{n_1} + \frac{s_2^2}{n_2}}},
\end{equation}

This ensured that the stratified sampling strategy did not introduce distributional bias, thereby supporting valid model generalization and downstream inference. Clinically, it confirmed that both sets of patients were comparable at baseline, so that observed differences in predicted renal outcomes could be attributed to model signals rather than sampling artifacts.

To assess the individual predictive utility of each feature, an ablation analysis was conducted by iteratively removing one variable at a time from the final model and retraining a logistic regression classifier\cite{Chen2025.05.11.25327405}. The impact of removal was measured by changes in AUROC, offering direct insight into each variable’s marginal contribution. This analysis helped highlight which physiologic and laboratory factors were most strongly associated with the risk of vancomycin-related renal impairment, operationalized as significant post-administration creatinine elevation.

Formally, the ablation effect \(\Delta AUC(x_i)\) for a given feature \(x_i\) was computed as:

\begin{equation}
\Delta AUC(x_i) = AUC_{\text{full}} - AUC_{-x_i}
\end{equation}

where \(AUC_{\text{full}}\) denotes the AUROC of the complete model including all features, and \(AUC_{-x_i}\) denotes the AUROC after removing feature \(x_i\). A larger \(\Delta AUC(x_i)\) indicates greater marginal importance of the feature in predicting renal risk.

To characterize non-linear associations and clinically relevant thresholds, we applied Accumulated Local Effects for top-ranked continuous variables. ALE curves provide unbiased estimates of a feature’s local impact on model predictions while addressing multicollinearity. Formally, the ALE for a given feature \(x_j\) is defined as:

\begin{equation}
ALE_j(z) = \int_{z_0}^{z} \mathbb{E}_{X_{\setminus j}} \left[ \frac{\partial f(X)}{\partial x_j} \Big| x_j = s \right] ds,
\end{equation}

These visualizations revealed interpretable patterns—such as saturation effects in phosphate and U-shaped trends in magnesium—that aligned with known renal physiology, particularly in the context of drug-induced tubular stress and hemodynamic injury. These insights help clinicians define physiologic thresholds where risk escalates sharply, enabling more timely monitoring or adjustment of nephrotoxic therapies.

To enhance patient-level interpretability and clinical auditability, we employed SHAP to decompose predictions into additive feature contributions
\cite{yang2023explainable}. Each SHAP value \(\phi_i\) represents the marginal contribution of a feature relative to all possible feature coalitions:

\begin{equation}
\phi_i = \sum_{S \subseteq F \setminus \{x_i\}} \frac{|S|! (|F|-|S|-1)!}{|F|!} \left[f(S \cup \{x_i\}) - f(S)\right],
\end{equation}

This allowed us to generate both global importance rankings and patient-specific explanation plots, improving model transparency and facilitating clinical trust. From a medical standpoint, SHAP enables clinicians to trace a patient’s predicted renal risk back to underlying contributing features—e.g., elevated lactate or low platelet count—offering rationale for interventions and supporting explainable AI in nephrotoxic drug management.

Finally, to quantify uncertainty in individual-level predictions, we implemented Bayesian posterior sampling using the DREAM algorithm. Unlike point estimates, this method generates a full posterior distribution for the predicted probability of vancomycin-associated creatinine elevation, allowing explicit quantification of prediction uncertainty.

The posterior predictive distribution was estimated as:

\begin{equation}
p(y \mid X) = \frac{1}{N \times C} \sum_{i=1}^{C} \sum_{j=1}^{N} p(y \mid \theta_{ij}, X)
\end{equation}

where \(N\) is the number of iterations per chain and \(C\) is the number of parallel chains. Each \(\theta_{ij}\) represents sampled model parameters from chain \(i\) at iteration \(j\). This multi-chain, multi-iteration approach improves parameter space exploration and sampling robustness.

DREAM is particularly suited for clinical prediction due to its adaptive proposal mechanism and efficient sampling in complex, high-dimensional ICU data. Unlike basic MCMC, DREAM dynamically adjusts its sampling strategy, ensuring more reliable posterior estimation without model retraining.

In ICU practice, this uncertainty-aware prediction supports patient-specific decisions. For example, a high predicted risk with a narrow credible interval may prompt early intervention, while wide intervals may suggest close monitoring instead of immediate treatment changes. This enables more personalized and balanced renal risk management, especially for vancomycin-treated patients.

In summary, Bayesian posterior sampling with DREAM provides an efficient, interpretable way to communicate model uncertainty, enhancing the practical value of creatinine elevation risk prediction in real-time ICU settings.

Together, these statistical analyses strengthened the model’s validity, interpretability, and real-world relevance. They ensured that our framework is not only predictive but also transparent and clinically actionable for forecasting early creatinine elevation and renal risk in ICU patients receiving vancomycin.

% Results and Discussion can be combined.
\section*{Results}
\subsection*{Cohort Characteristics and Statistical Comparison}

The study cohort consists of critically ill patients who received vancomycin during their ICU stay. This study specifically focuses on identifying patient profiles that are more prone to developing significant creatinine elevation following vancomycin exposure. Although the creatinine changes observed meet the diagnostic thresholds typically used for AKI, the research target is the creatinine response associated with vancomycin, not the broader clinical syndrome of AKI. This distinction emphasizes the interest in vancomycin-induced renal effects rather than all-cause AKI.

The dataset was randomly divided using stratified sampling into a training set (70\%) and a test set (30\%), ensuring that the distribution of creatinine elevation events was balanced across both subsets. This partitioning minimizes sampling bias and enhances the reliability and generalizability of the model.

As shown in Table~\ref{tab:cohort_comparison_results}, none of the 19 clinical variables demonstrated statistically significant differences ($p > 0.05$) between the training and test sets, confirming the internal consistency and comparability of the two cohorts.

Table~\ref{tab:cohort_comparison_results_1} presents the baseline characteristics of the creatinine elevation and non-elevation groups. Patients with creatinine elevation exhibited higher levels of phosphate, lactate, PTT, anion gap, AST, and total bilirubin, along with lower platelet counts, more negative arterial base excess, and reduced Richmond-RAS scores. These differences suggest a combination of metabolic, coagulative, and neurologic disturbances. Additionally, the creatinine elevation group tended to be older and had higher Charlson Comorbidity Index scores, indicating a greater chronic disease burden.

These findings are consistent with known mechanisms of vancomycin-associated nephrotoxicity, including oxidative stress, tubular injury, and impaired renal perfusion. By focusing specifically on creatinine elevation linked to vancomycin use, this study offers a targeted perspective on medication-associated renal risk, rather than the generalized context of ICU-related AKI. The internal consistency of the dataset, along with the biological plausibility of the identified predictors, supports the clinical relevance and reliability of the proposed model in assessing vancomycin-associated creatinine elevation risk.

\begin{table}[H]
\caption{\textbf{T-test Comparison of Feature Distributions between Training and Test Sets.}}
\label{tab:cohort_comparison_results}
\small
\renewcommand{\arraystretch}{1.2}
\rowcolors{2}{white}{white}
\begin{tabularx}{\textwidth}{l|l|X|X|l}
\hline
\rowcolor[HTML]{f7e1d7}
\textbf{Feature} & \textbf{Unit} & \textbf{Training Set} & \textbf{Test Set} & \textbf{P-value} \\ \hline
Phosphate & mg/dL & 3.61 (1.19) & 3.62 (1.20) & 0.555 \\ \hline
APS III & -- & 51.99 (22.52) & 51.63 (22.05) & 0.456 \\ \hline
Magnesium & mg/dL & 2.09 (0.30) & 2.09 (0.30) & 0.958 \\ \hline
Lactate & mmol/L & 2.20 (1.46) & 2.19 (1.41) & 0.654 \\ \hline
PTT & sec & 40.05 (16.09) & 40.66 (16.81) & 0.089 \\ \hline
Anion Gap & mmol/L & 14.17 (3.59) & 14.06 (3.47) & 0.154 \\ \hline
Platelet Count & $\times$10\textsuperscript{3}/µL & 197.33 (108.19) & 196.41 (107.81) & 0.690 \\ \hline
Arterial Base Excess & mmol/L & -1.33 (3.72) & -1.27 (3.86) & 0.413 \\ \hline
Richmond-RAS Scale & -- & -1.31 (1.35) & -1.31 (1.39) & 0.923 \\ \hline
White Blood Cells & $\times$10\textsuperscript{3}/µL & 13.03 (9.21) & 12.63 (7.63) & 0.023 \\ \hline
Glucose & mg/dL & 144.82 (49.59) & 145.62 (48.87) & 0.447 \\ \hline
Mean Airway Pressure & cmH\textsubscript{2}O & 10.50 (3.28) & 10.52 (3.29) & 0.821 \\ \hline
Total Bilirubin & mg/dL & 2.36 (4.23) & 2.26 (4.11) & 0.287 \\ \hline
AST & U/L & 319.67 (1000.05) & 281.29 (743.46) & 0.061 \\ \hline
Age & years & 61.27 (14.28) & 61.16 (14.31) & 0.701 \\ \hline
Braden Mobility & score & 2.39 (0.59) & 2.40 (0.59) & 0.580 \\ \hline
ED Duration & hours & 3.47 (4.60) & 3.42 (4.27) & 0.582 \\ \hline
Charlson Comorbidity Index & score & 4.61 (2.76) & 4.66 (2.81) & 0.473 \\ \hline
Arterial Line & binary & 0.61 (0.49) & 0.60 (0.49) & 0.219 \\ \hline
\end{tabularx}
\begin{flushleft}
\textit{Note}: This table summarizes statistical comparisons between the training and test cohorts. Continuous variables are expressed as mean (standard deviation). P-values are derived from two-sided t-tests, with significance set at $p<0.05$.
\end{flushleft}
\end{table}

\begin{table}[H]
\caption{\textbf{T-test Comparison of Feature Distributions Between Non-Elevation and Elevation.}}
\label{tab:cohort_comparison_results_1}
\small
\renewcommand{\arraystretch}{1.2}
\rowcolors{2}{white}{white}
\begin{tabularx}{\textwidth}{l|l|X|X|l}
\hline
\rowcolor[HTML]{f7e1d7}
\textbf{Feature} & \textbf{Unit} & \textbf{Non-Elevation Set} & \textbf{Elevation Set} & \textbf{P-value} \\ \hline
Phosphate & mg/dL & 3.40 (1.07) & 4.13 (1.31) & < 0.001 \\ \hline
APS III & -- & 48.66 (21.05) & 60.44 (23.89) & < 0.001 \\ \hline
Magnesium & mg/dL & 2.06 (0.28) & 2.16 (0.32) & < 0.001 \\ \hline
Lactate & mmol/L & 2.05 (1.26) & 2.58 (1.83) & < 0.001 \\ \hline
PTT & sec & 38.65 (15.22) & 43.62 (17.64) & < 0.001 \\ \hline
Anion Gap & mmol/L & 13.75 (3.34) & 15.23 (3.96) & < 0.001 \\ \hline
Platelet Count & $\times$10\textsuperscript{3}/µL & 202.85 (106.86) & 183.31 (110.30) & < 0.001 \\ \hline
Arterial Base Excess & mmol/L & -0.99 (3.57) & -2.21 (3.96) & < 0.001 \\ \hline
Richmond-RAS Scale & -- & -1.17 (1.26) & -1.65 (1.50) & < 0.001 \\ \hline
White Blood Cells & $\times$10\textsuperscript{3}/µL & 12.83 (9.72) & 13.52 (7.72) & 0.001 \\ \hline
Glucose & mg/dL & 142.96 (48.24) & 149.55 (52.59) & < 0.001 \\ \hline
Mean Airway Pressure & cmH\textsubscript{2}O & 10.24 (3.07) & 11.17 (3.69) & < 0.001 \\ \hline
Total Bilirubin & mg/dL & 1.99 (3.15) & 3.28 (6.08) & < 0.001 \\ \hline
AST & U/L & 250.16 (758.82) & 496.48 (1427.11) & < 0.001 \\ \hline
Age & years & 60.80 (14.53) & 62.47 (13.55) & < 0.001 \\ \hline
Braden Mobility & score & 2.44 (0.58) & 2.25 (0.59) & < 0.001 \\ \hline
ED Duration & hours & 3.51 (4.09) & 3.36 (5.68) & 0.263 \\ \hline
Charlson Comorbidity Index & score & 4.37 (2.72) & 5.24 (2.75) & < 0.001 \\ \hline
Arterial Line & binary & 0.57 (0.50) & 0.71 (0.45) & < 0.001 \\ \hline
\end{tabularx}
\begin{flushleft}
\textit{Note}: This table compares patients with and without vancomycin-associated creatinine elevation. Continuous features are presented as mean (standard deviation). P-values were calculated using two-sided t-tests with a significance threshold of $p<0.05$. The creatinine elevation events included in this study meet established diagnostic thresholds commonly used for AKI but specifically represent vancomycin-associated renal response.
\end{flushleft}
\end{table}

\subsection*{Feature Contribution Analysis}

\begin{figure}[H]
    \centering
    \includegraphics[width=1\linewidth]{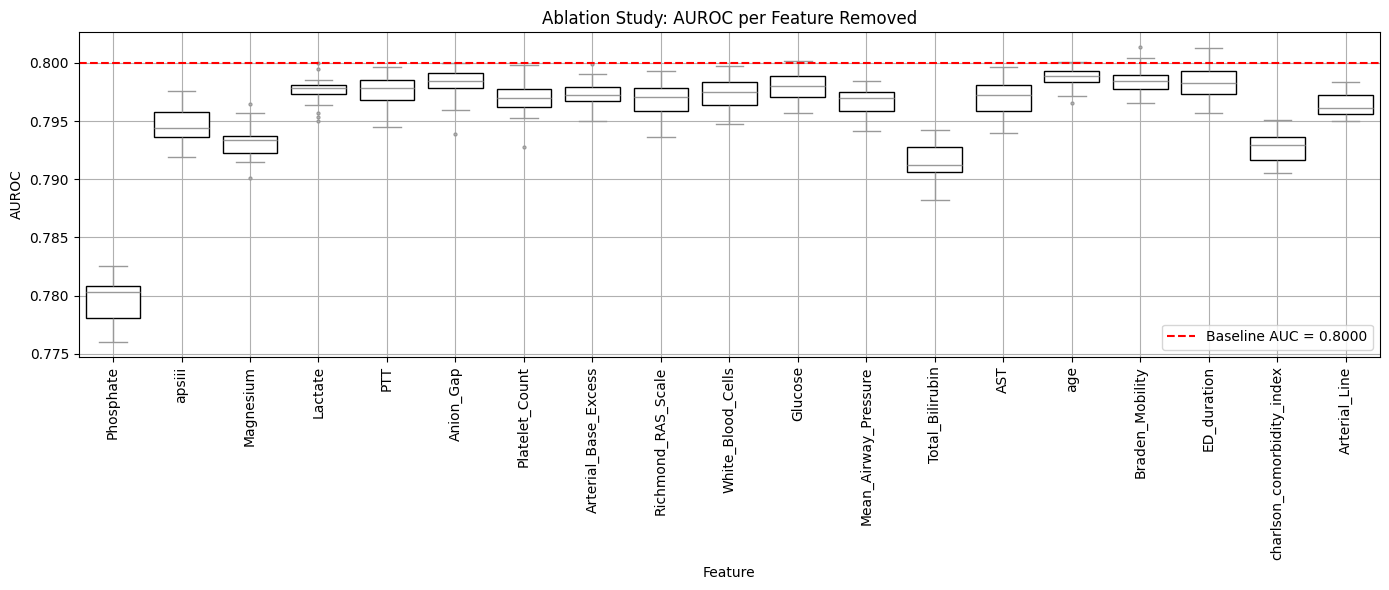}
    \caption{\textbf{Impact of Feature Removal on LR Model Performance.}}
    \label{fig:ablation_analysis}
\end{figure}

To evaluate individual feature contributions for predicting vancomycin-associated creatinine elevation in critically ill patients, we conducted an ablation analysis. As illustrated in Figure~\ref{fig:ablation_analysis}, each feature was systematically removed, and a logistic regression classifier was retrained using bootstrap sampling to assess its marginal impact on AUROC. The red dashed line represents the baseline AUROC (0.800) achieved when all features were included. This analysis aimed to quantify the predictive utility of each variable independently of specific model architectures, providing information on which clinical factors most strongly contribute to the evaluation of nephrotoxicity risk.

Notably, the exclusion of phosphate led to the most substantial decline in model performance (AUROC dropping to approximately 0.780), indicating that phosphate plays the dominant role in identifying patients at higher risk of creatinine elevation following vancomycin exposure. This is consistent with clinical observations that abnormal phosphate levels may reflect early renal stress or tubular dysfunction, which are critical in the development of drug-associated nephrotoxicity.

Other features with considerable impact included APSIII and magnesium, which showed notable performance decreases when removed (AUROC approximately 0.793 and 0.792 respectively), highlighting the importance of acute illness severity and electrolyte balance in nephrotoxicity risk assessment. Charlson Comorbidity Index and arterial line presence also demonstrated moderate contributions to predictive performance. In contrast, features such as ED duration, anion gap, and several laboratory values showed minimal impact when excluded, suggesting these variables provide limited unique predictive information in this clinical context.

It is important to emphasize that this ablation analysis serves a different purpose than traditional model ablation studies. Rather than evaluating model robustness, this analysis specifically quantifies individual feature contributions using logistic regression as a standardized baseline classifier. This approach provides model-agnostic insights into feature utility that may differ from contributions in more complex architectures, as non-linear interactions and feature substitution effects are not captured by the linear baseline.

Overall, the ablation results demonstrate clear hierarchical importance among clinical variables, with phosphate, acute illness severity, and key electrolyte markers providing the strongest individual contributions to vancomycin nephrotoxicity prediction. These findings support the biological plausibility of the feature set and provide valuable insights for clinical risk factor prioritization.

\subsection*{Model Performance on Creatinine
Elevation Risk Prediction}

\begin{figure}[H]
\centering
\includegraphics[width=0.95\linewidth]{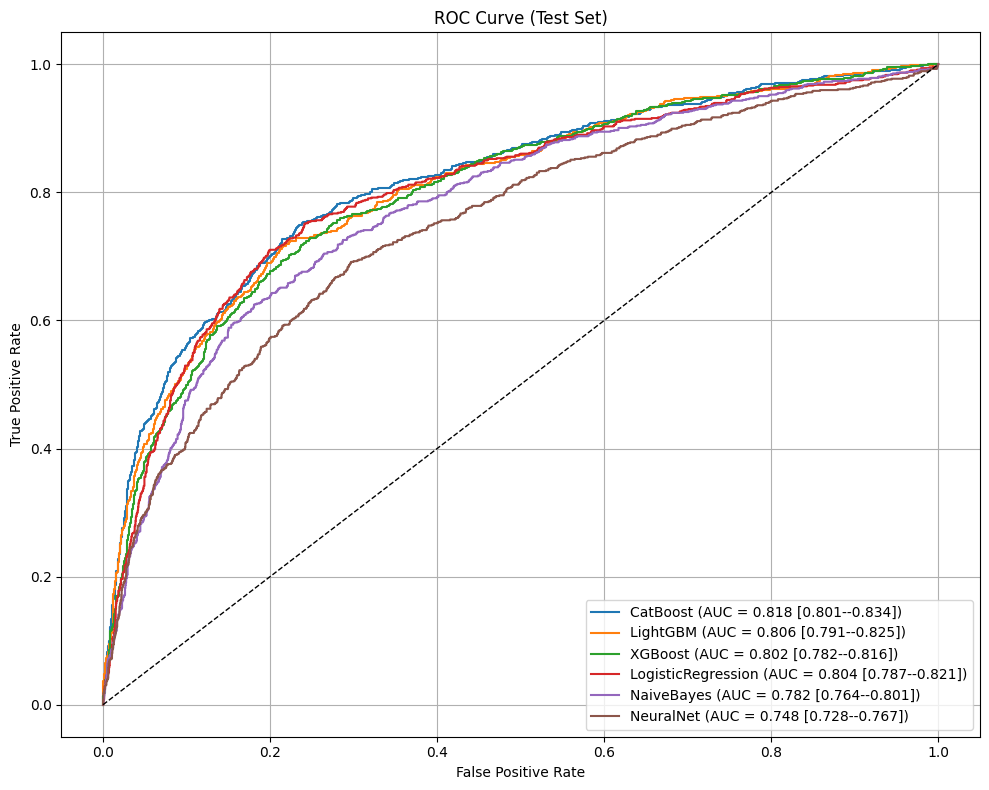}
\caption{\textbf{AUROC Curves for Model Performance in the Test Set.}}
\label{fig:roc_test}
\end{figure}

To evaluate the capacity of different algorithms to predict vancomycin-associated creatinine elevation among ICU patients, we tested six widely used machine learning models. Performance metrics on the test set—including AUROC, sensitivity, specificity, F1-score, and predictive values—are summarized in Table~\ref{tab: Results of the Test Set}. ROC curves for the test set are illustrated in Figure~\ref{fig:roc_test}.

Our dataset exhibited moderate class imbalance with 28.2\% nephrotoxicity cases (2,903 of 10,288 patients). To ensure fair comparison across algorithms, sensitivity was manually fixed at 0.800 for all models, allowing direct evaluation of specificity and precision trade-offs at this clinically relevant threshold.

Among the models evaluated, CatBoost achieved the highest AUROC of 0.818 (95\% CI: 0.801–0.834), indicating strong discriminatory ability. It also delivered the best overall accuracy (0.714), highest F1-score (0.605), and maintained a solid specificity of 0.681 at the fixed sensitivity threshold, ensuring the model captures most high-risk patients without overwhelming clinicians with false alarms. Critically, CatBoost's performance demonstrates genuine predictive skill beyond baseline rates: its PPV of 0.486 represents a 72\% improvement over the 0.282 baseline prevalence, while the NPV of 0.900 substantially exceeds the 0.718 rate a naive "always safe" classifier would achieve. This translates to correctly identifying an additional 182 low-risk patients per 1,000 predictions beyond chance alone, confirming clinically meaningful risk stratification that substantially exceeds baseline performance expectations.

From a clinical perspective, this level of performance is particularly valuable in real-world ICU settings where vancomycin is commonly used to treat severe infections but carries well-documented nephrotoxic potential. Even a modest rise in creatinine can signal the early stages of renal injury in ICU patients, where rapid clinical deterioration is possible. In practice, this model can support timely risk stratification by identifying patients who may benefit from intensified renal monitoring, dose adjustment, or consideration of alternative therapies. For example, correctly flagging 80\% of future creatinine elevation cases while still safely ruling out nearly 70\% of low-risk patients provides actionable guidance that can directly inform bedside decisions.

The high NPV (0.900) is particularly valuable for clinical decision-making, enabling two complementary treatment strategies: confidently continuing vancomycin therapy in patients predicted as low-risk while implementing enhanced monitoring protocols for high-risk patients. This dual approach helps avoid both unnecessary treatment interruptions in safe patients and delayed intervention in vulnerable patients, optimizing antimicrobial stewardship without compromising patient safety.

CatBoost was ultimately selected as the final model not only for its superior test performance but also for its robustness to missing data and its ability to handle heterogeneous ICU feature sets. Its interpretability is strengthened by its reliance on physiologically meaningful predictors, such as phosphate, bilirubin, and comorbidity burden, which are strongly associated with renal stress mechanisms linked to vancomycin use. This ensures that the model’s outputs are transparent and clinically intuitive, even for readers without a background in machine learning.

In summary, by prioritizing high sensitivity and carefully balancing specificity, this framework provides a practical, explainable tool to support early detection of vancomycin-associated creatinine elevation and renal injury risk in ICU patients. It offers meaningful, real-time clinical support without compromising interpretability or safety.

\begin{table}[H]
\renewcommand{\arraystretch}{1.2}
\centering
\caption{\textbf{Performance Comparison of Different Models in the Test Set.}}
\resizebox{\textwidth}{!}{
\begin{tabular}{l|l|l|l|l|l|l|l}
\hline
\rowcolor[HTML]{f7e1d7}
\textbf{Model} & \textbf{AUC (95\% CI)} & \textbf{Accuracy} & \textbf{F1-score} & \textbf{Sensitivity} & \textbf{Specificity} & \textbf{PPV} & \textbf{NPV} \\ \hline
\rowcolor[HTML]{a8dadc}
CatBoost & 0.818 (0.801--0.834) & 0.714 & 0.605 & 0.800 & 0.681 & 0.486 & 0.900 \\
LightGBM & 0.806 (0.791--0.825) & 0.690 & 0.587 & 0.800 & 0.648 & 0.462 & 0.896 \\
XGBoost & 0.802 (0.782--0.816) & 0.677 & 0.576 & 0.800 & 0.631 & 0.450 & 0.893 \\
LogisticRegression & 0.804 (0.787--0.821) & 0.694 & 0.589 & 0.800 & 0.653 & 0.466 & 0.897 \\
NaiveBayes & 0.782 (0.764--0.801) & 0.647 & 0.554 & 0.800 & 0.589 & 0.423 & 0.886 \\
NeuralNet & 0.748 (0.728--0.767) & 0.596 & 0.521 & 0.800 & 0.519 & 0.386 & 0.873 \\ \hline
\end{tabular}
}
\label{tab: Results of the Test Set}
\end{table}

\subsection*{SHAP analysis and feature attribution}

\begin{figure}[H]
    \centering
    \includegraphics[width=0.95\linewidth]{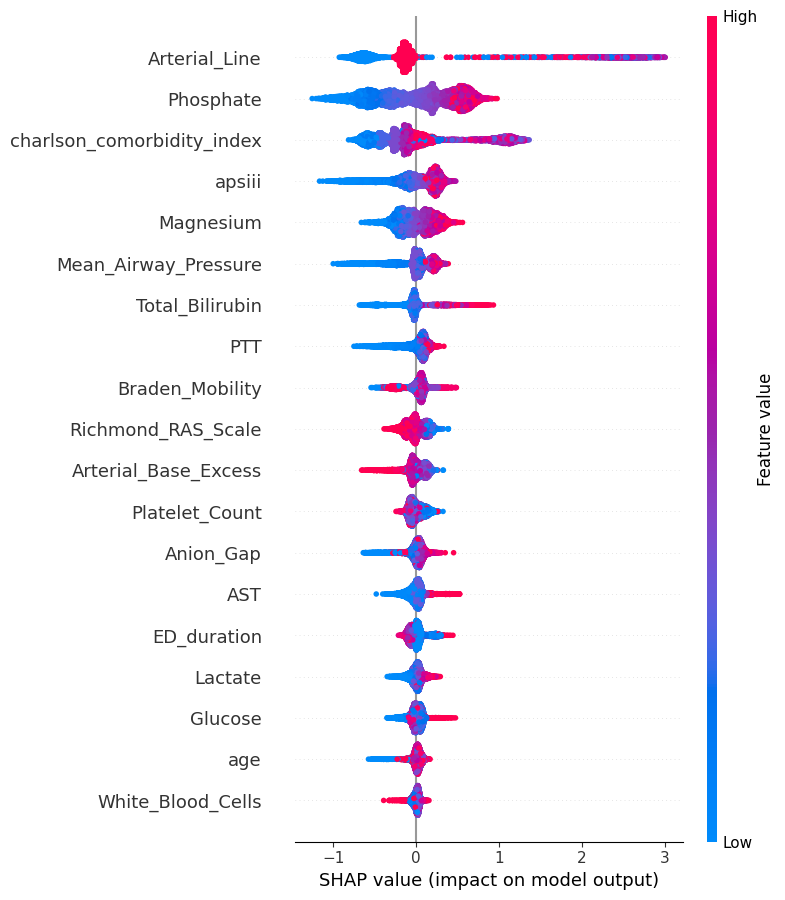}
    \caption{\textbf{SHAP summary plot showing feature contributions to predicted creatinine elevation in vancomycin patients.}}
    \label{fig:shap_summary}
\end{figure}

To interpret the contribution of individual variables in predicting vancomycin-associated creatinine elevation among ICU patients, SHAP analysis was applied to the CatBoost model. Figure~\ref{fig:shap_summary} presents a SHAP summary plot, where each point represents an individual patient. The x-axis shows the SHAP value, which reflects the degree to which a feature influences the model's prediction for that patient. Colors indicate the feature value: red points correspond to high values, and blue points represent low values.

Phosphate was the most influential predictor, with high phosphate levels (red points) consistently positioned on the right side of the plot, indicating that elevated phosphate levels substantially increase the predicted risk of creatinine elevation. This pattern is clinically plausible, as phosphate retention may signal impaired renal handling and early kidney stress. 

Total bilirubin and magnesium also ranked among the top contributors. For these features, red and blue points are more evenly spread across the SHAP axis, suggesting complex and potentially non-linear relationships. For example, the impact of magnesium appears bidirectional, where both low and high magnesium levels may influence creatinine trajectories, which is reflected in red and blue points appearing on both sides of the SHAP axis.

Charlson comorbidity index and APS III also demonstrated significant impact. Higher scores (red points) typically increased predicted risk, reflecting the influence of chronic disease burden and acute illness severity on renal vulnerability. The more concentrated distribution of red points on the right side for these variables suggests a relatively straightforward, monotonic relationship.

The colored point distribution across the SHAP axis offers insight into both the importance and the detailed effect patterns of each variable. Features where red points consistently shift predictions to the right imply a clear risk-enhancing role, while mixed distributions suggest subtler, context-dependent effects. The model’s reliance on physiologically meaningful predictors, as reflected in these SHAP patterns, supports its clinical interpretability and enhances its potential for integration into real-world ICU risk management, where early identification of vancomycin-related renal injury risk can inform timely monitoring and intervention.

\subsection*{ALE analysis and clinical interpretability}

\begin{figure}[H]
    \centering
    \includegraphics[width=0.85\linewidth]{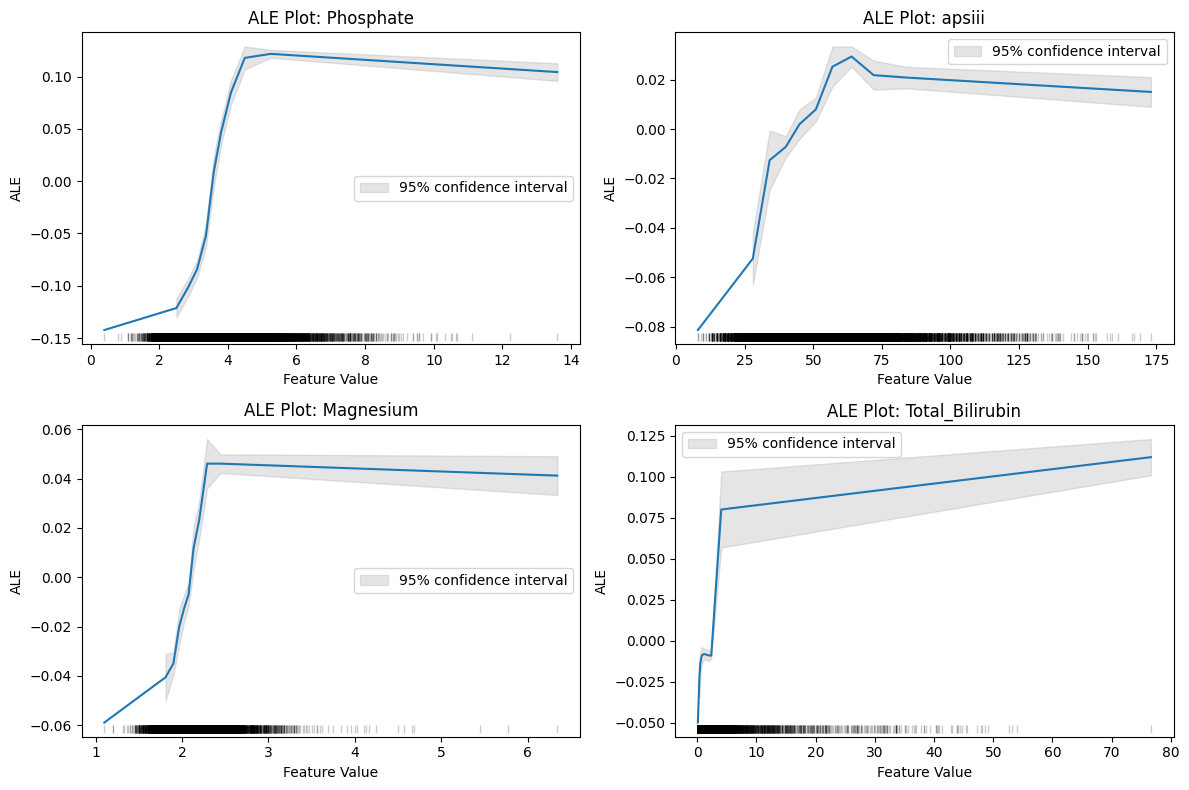}
    \caption{\textbf{ALE plots for top features in vancomycin ICU patients.}}
    \label{fig:ale_analysis}
\end{figure}

To further investigate the local interpretability and clinical plausibility of our CatBoost model in predicting vancomycin-associated creatinine elevation among ICU patients, we conducted an ALE analysis focusing on four high-impact features: phosphate, APSIII, magnesium, and total bilirubin. The ALE plots, shown in Figure~\ref{fig:ale_analysis}, illustrate the marginal effect of each variable on the model's output while accounting for the influence of other features.

Phosphate exhibited a steep positive ALE effect between values of 2 and 5 mg/dL, after which the curve plateaus and slightly decreases. This suggests that increasing phosphate levels strongly raise the predicted risk of creatinine elevation up to a certain threshold, beyond which the effect saturates. The sharp rise at lower levels aligns with clinical understanding that even moderate phosphate elevation may reflect early renal stress or impaired clearance, especially in vancomycin-treated patients.

For APSIII, a widely used severity score, the ALE curve shows a continuous upward trend up to approximately 70, indicating that patients with more severe physiological disturbances are more likely to experience creatinine elevation following vancomycin exposure. The flattening of the curve at higher APSIII scores likely reflects the model’s conservative adjustments in data-sparse regions, which is a desirable trait for maintaining prediction stability.

Magnesium demonstrated a non-linear effect: the ALE sharply increases from 1.5 to approximately 2.5 mEq/L, then stabilizes or slightly decreases. This pattern may suggest that both low and excessively high magnesium levels are clinically relevant, but the model is most sensitive to moderate elevations in magnesium, which may be linked to systemic imbalance or renal susceptibility.

Total bilirubin’s ALE plot shows a rapid positive effect starting at low levels, which continues to rise gradually. This indicates that elevated bilirubin consistently increases the predicted risk, possibly reflecting the impact of hepatic-renal interactions or systemic inflammation that can exacerbate renal stress in critically ill patients.

Overall, the ALE analysis confirms that the model’s risk estimations are biologically coherent and sensitive to clinically relevant ranges of key predictors. The smooth, interpretable curves indicate that the model does not overfit to outliers and responds appropriately to gradual physiological changes. These results support the model's potential for real-time risk stratification in ICU settings, where early identification of vancomycin-associated creatinine elevation can guide timely monitoring and intervention to mitigate renal injury risk.

\subsection*{Posterior Distribution and Prediction of Vancomycin-Related Creatinine Elevation}

To incorporate uncertainty into the prediction of vancomycin-associated creatinine elevation among ICU patients, we applied the DREAM algorithm to the trained CatBoost model. Unlike traditional deterministic predictions, this approach generates a posterior distribution over the predicted probability, allowing clinicians to assess not only the most likely risk estimate but also the associated confidence range for individual patients. This is particularly valuable in ICU settings, where misjudging renal risk can lead to either delayed intervention or unnecessary treatment changes.

In this study, we set the number of iterations to 2000 and used 38 parallel chains, following the recommended practice of employing at least twice the number of model parameters (19 variables) to ensure sufficient posterior exploration. This configuration allows the DREAM algorithm to efficiently traverse the parameter space, reduce the risk of local convergence, and produce stable, reliable posterior distributions. Adequate iterations and chains are essential to capture uncertainty accurately, particularly in high-stakes ICU risk prediction tasks.

\begin{figure}[H]
    \centering
    \includegraphics[width=0.85\linewidth]{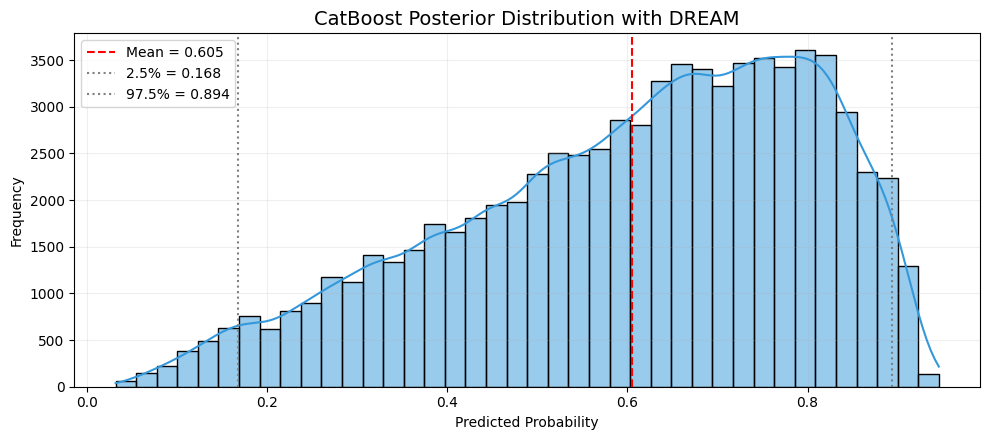}
    \caption{\textbf{Posterior distribution of vancomycin-associated creatinine elevation risk for a high-risk ICU patient.}}
    \label{fig:uq_catboost}
\end{figure}

As shown in Figure~\ref{fig:uq_catboost}, the posterior distribution for a representative high-risk patient is skewed toward higher predicted probabilities, with a mean risk of 60.5\% and a 95\% credible interval ranging from 16.8\% to 89.4\%. In comparison to the overall cohort creatinine elevation rate of 28.22\%, this patient demonstrates substantially increased risk. The wide credible interval highlights clinical uncertainty, indicating that while the model identifies this patient as high-risk, variability in clinical features could significantly influence the actual outcome.

The high-risk patient profile used in the posterior simulation was constructed based on the creatinine elevation subgroup described in Table~\ref{tab:cohort_comparison_results_1}, which includes elevated phosphate, bilirubin, magnesium, and higher Charlson comorbidity index scores. This ensures that the sampling process is grounded in real-world ICU scenarios and reflects clinically plausible risk patterns for vancomycin-associated renal injury, rather than relying on theoretical or averaged inputs.

This probabilistic framework provides meaningful clinical nuance. Two patients may have similar mean predicted risks but differ in the width of their credible intervals—one presenting with high certainty and another with considerable uncertainty. For patients with wide intervals, clinicians may choose to prioritize enhanced monitoring over immediate medication adjustments, recognizing the potential variability in risk trajectories.

Importantly, DREAM operates without retraining the CatBoost model. It conditions posterior sampling on prior distributions derived from observed creatinine elevation cases, enabling computationally efficient uncertainty quantification that can be applied in real-time. This makes the approach practical for bedside decision support.

In summary, integrating uncertainty-aware prediction with CatBoost and DREAM offers a clinically interpretable and statistically robust framework to assess the risk of vancomycin-associated creatinine elevation in ICU patients. This enhances clinical confidence in machine learning-assisted decisions, particularly in managing nephrotoxic therapy and optimizing renal risk surveillance.

\section*{Discussion}
\subsection*{Summary of Existing Model Compilation}
In this study, we developed a machine learning pipeline using the MIMIC-IV critical care database to predict vancomycin-associated renal injury in ICU patients, defined by significant elevations in serum creatinine. Our approach encompassed data extraction of ICU stays where patients received vancomycin, rigorous preprocessing and feature engineering, and the training/evaluation of multiple candidate algorithms. Notably, an ensemble of both linear and non-linear models (including logistic regression, random forests, gradient boosting variants, and neural networks) was explored. Among these, the CatBoost gradient boosting model emerged as the top performer, yielding the highest discrimination for predicting nephrotoxic injury. CatBoost’s superior performance can be attributed to its ability to handle heterogeneous clinical features (including categorical variables and missing values) and capture complex non-linear interactions without extensive tuning. Using nested cross-validation, CatBoost achieved the best overall accuracy and calibration, outperforming other models on metrics such as AUC and F1-score. The final CatBoost model demonstrated robust predictive ability for early identification of patients at risk of vancomycin-induced acute kidney injury (AKI). We also examined feature importance to interpret the model: key predictors of vancomycin-associated creatinine rise included baseline renal function (initial creatinine), vancomycin exposure factors (e.g. dosing intensity or duration), patient comorbidities, and concurrent nephrotoxic treatments. These findings underscore that a data-driven ML approach can synthesize many risk factors into an effective predictor, potentially enabling early warnings for impending renal impairment during vancomycin therapy.

\subsection*{Comparison with Prior Studies}
Our results align with and extend the observations of prior clinical studies on vancomycin’s nephrotoxic effects. It is well established that vancomycin has a narrow therapeutic window and can precipitate kidney injury, especially in critically ill patients; reported incidences of AKI in ICU patients on vancomycin range up to 40\%~\cite{zamoner2022}. Earlier investigations primarily focused on identifying risk factors for vancomycin-induced nephrotoxicity. For instance, high vancomycin trough concentrations (15–20\,mg/L) have consistently been associated with increased rates of creatinine elevation and AKI. van Hal et al.~\cite{vanhal2013} conducted a meta-analysis and found that maintaining trough levels in the 15–20\,mg/L range significantly heightens nephrotoxicity risk compared to lower targets. Likewise, Bosso et al.~\cite{bosso2011} reported a clear relationship between elevated vancomycin troughs and subsequent acute renal dysfunction in a prospective multicenter trial. These studies support the paradigm that aggressive vancomycin dosing can be harmful to the kidneys, corroborating our model’s inclusion of vancomycin exposure metrics as important predictors.

Beyond drug levels, patient-specific factors identified previously are in line with our findings. Age and critical illness severity, for example, are known to modulate vancomycin’s nephrotoxic risk. One study observed that patients over 80 years old experienced higher rates of vancomycin-related renal injury~\cite{li2022}. Concomitant nephrotoxic agents also play a role: the combination of vancomycin with piperacillin–tazobactam has been associated with a significantly higher incidence of AKI than vancomycin alone~\cite{blair2021}. Our model implicitly captured such effects—for example, the presence of nephrotoxin co-administration featured among the most influential predictors contributing to risk stratification.

Notably, few earlier efforts have used machine learning for this problem. A recent study by Aghamirzaei et al.~\cite{aghamirzaei2025} developed a stacking ensemble model for vancomycin-induced AKI prediction in 314 ICU patients, reporting an impressive AUC of 0.94. Their model highlighted variables like serum creatinine trends and glucose variability. Our study builds on this with a substantially larger cohort from MIMIC-IV and demonstrates that even a single advanced algorithm (CatBoost) can achieve high predictive performance. In contrast to rule-based clinical scoring or traditional regression, our ML approach offers improved sensitivity for complex, non-linear interactions and patient heterogeneity.

A recent study by Bao et al. published in \textit{Frontiers in Pharmacology}~\cite{bao2024nomogram} also explored vancomycin-associated nephrotoxicity using the MIMIC-IV database. However, their method employed AKI diagnosis labels without access to exact onset times, a limitation stemming from MIMIC-IV's lack of time-stamped AKI events. This hinders the ability to confirm that AKI occurred after vancomycin administration. In contrast, our study uses time-stamped serum creatinine trajectories to define kidney injury, ensuring that elevated creatinine levels follow vancomycin initiation. This temporal precision enhances causal interpretability and clinical relevance. Moreover, our focus on routine creatinine monitoring supports real-time risk prediction at the bedside—an advantage over AKI-label-based approaches that lack precise timing and are less actionable.

Additionally, pharmacologic monitoring supports the value of early prediction. Zamoner et al.~\cite{zamoner2022} found that an abnormal vancomycin serum level could predict impending AKI roughly 48 hours before diagnosis in septic ICU patients. Such findings resonate with our goal of early detection. By deploying a predictive model that continuously ingests routine EHR data—including labs, vitals, and drug administration—our work contributes a practical approach for real-time clinical integration. In summary, whereas prior studies established the risk factors and incidence of vancomycin nephrotoxicity, our model integrates those insights into a comprehensive, interpretable, and prospective prediction tool, enabling earlier intervention and improved patient safety.

\subsection*{Limitations and Future Works}

This study has several important limitations. First, our retrospective design using MIMIC-IV data from a single academic medical center may limit generalizability to other ICU populations with different patient demographics, clinical protocols, or resource constraints. The temporal span (2008-2019) may not reflect current clinical practices, potentially affecting model performance in contemporary settings.

Second, our nephrotoxicity definition relies solely on serum creatinine elevation, which is a delayed marker that may miss subclinical renal injury or cases where elevation is masked by clinical factors. The absence of more sensitive biomarkers (NGAL, cystatin C) or functional assessments limits our ability to detect early nephrotoxicity.

Third, methodological constraints include potential imputation bias from missing data patterns, unmeasured confounding factors (genetic predisposition, subclinical kidney disease, undocumented nephrotoxic exposures), and feature selection limitations that may have excluded clinically relevant variables such as detailed vancomycin pharmacokinetics or hemodynamic parameters.

Fourth, while our CatBoost model achieved strong performance, its ensemble nature creates interpretability challenges for clinical adoption. The evaluation focused primarily on discrimination metrics without extensive exploration of calibration or clinical utility measures, and performance thresholds were chosen arbitrarily rather than through clinical consensus.

Finally, clinical implementation faces significant barriers including workflow integration challenges, regulatory validation requirements, and potential algorithmic bias across demographic groups. Our model requires real-time access to multiple data streams that may not be consistently available in all clinical settings.

Future work should prioritize multi-center external validation across diverse healthcare systems to establish broader generalizability. Prospective randomized controlled trials comparing model-guided versus standard monitoring strategies are essential to demonstrate clinical utility and cost-effectiveness using meaningful endpoints such as time to nephrotoxicity detection and clinical outcomes.

Methodological enhancements should include integration of advanced biomarkers when available, development of longitudinal time-series models to capture dynamic risk evolution, and incorporation of detailed vancomycin pharmacokinetic data for more mechanistically informed predictions. Advanced analytical approaches should employ causal inference techniques, systematic fairness evaluation across demographic subgroups, and federated learning approaches for collaborative model development while preserving data privacy.

Implementation research should investigate human-AI collaboration patterns, conduct comprehensive health economic evaluations, and examine organizational barriers to clinical adoption. The development of bias mitigation strategies and systematic evaluation of model performance across diverse patient populations will be crucial for ensuring equitable clinical application.

Future success depends on addressing these limitations through rigorous validation studies, enhanced methodological approaches, and careful attention to clinical implementation challenges. Collaborative efforts among data scientists, clinicians, regulatory bodies, and healthcare systems will be essential to ensure these tools meaningfully improve patient outcomes while maintaining safety and equity in clinical care.

\section*{Conclusion}

This study demonstrates that machine learning can successfully predict vancomycin-associated nephrotoxicity using routinely collected ICU data, achieving clinically meaningful performance while maintaining interpretability and supporting real-world implementation. The framework represents a significant step toward personalized, data-driven approaches to drug safety monitoring in critical care.
However, the transition from research demonstration to clinical implementation requires continued commitment to rigorous validation, careful attention to clinical workflow integration, and systematic consideration of ethical implications. The ultimate measure of success will not be algorithmic performance alone, but demonstrable improvement in patient outcomes, clinical efficiency, and healthcare equity.
By bridging the gap between retrospective risk identification and prospective, actionable prediction, this work underscores the transformative potential of interpretable machine learning for enhancing drug safety and optimizing clinical care in high-risk populations. The path forward demands collaborative efforts among researchers, clinicians, healthcare systems, and regulatory bodies to ensure that these technological advances translate into meaningful improvements in patient care and clinical outcomes.
Future research must balance methodological sophistication with clinical practicality, ensuring that the promise of precision medicine in drug safety monitoring becomes a reality that serves all patients effectively and equitably. The foundation established by this work provides a robust platform for continued innovation in machine learning applications for healthcare, with the ultimate goal of improving patient safety and clinical decision-making in critical care environments.

\section*{Acknowledgments}
J.F. conceptualized the study, developed the methodological framework, conducted the experiments, performed data analysis, and drafted the original manuscript. L.S. and S.C. contributed to data preprocessing, model development, and manuscript preparation. Y.S. and M.A. provided technical support for machine learning implementation and model validation. G.P., E.P., and K.A. provided critical feedback on the study design and interpretation of results. M.P. supervised the project, coordinated research efforts, and provided strategic guidance throughout the study. All authors reviewed and approved the final version of the manuscript.

The authors would also like to acknowledge the Laboratory for Computational Physiology at the Massachusetts Institute of Technology for maintaining the MIMIC-IV database.

\section*{Data Availability Statement}
The datasets used in this study are available through PhysioNet, a repository of freely-available medical research data, managed by the MIT Laboratory for Computational Physiology. The MIMIC-IV database can be accessed at PhysioNet \cite{johnson2020mimic} after completing the required training course and signing the data use agreement. Access requires registration and approval but is free for researchers. Due to the sensitive nature of clinical data, direct sharing of the processed datasets is not permitted under the data use agreement, but our analysis code and methodology are fully described to enable reproduction of results.
% Either type in your references using
% \begin{thebibliography}{}
% Text
% \end{thebibliography}
%
% or
%
% Compile your BiBTeX database using our plos2015.bst
% style file and paste the contents of your .bbl file
% here. See http://journals.plos.org/plosone/s/latex for 

\bibliography{referrences}

\end{document}